\pdfoutput=1
\documentclass[a4paper, twoside]{article}

\usepackage{epsfig}
\usepackage{subcaption}
\usepackage{calc}
\usepackage{amssymb}
\usepackage{amstext}
\usepackage{multicol}
\usepackage{pslatex}
\usepackage{apalike}
\usepackage{hyperref}
\usepackage{tabto}    
\usepackage[bottom]{footmisc}
\usepackage{SCITEPRESS}     % Please add other packages that you may need BEFORE the SCITEPRESS.sty package.

\newif\ifanonymous
\anonymousfalse

\begin{document}

\title{Sentiment-based Engagement Strategies for intuitive Human-Robot Interaction}

\ifanonymous
    \author{\authorname{Name \orcidAuthor{0000-0000-0000-0000},Name\orcidAuthor{0000-0000-0000-0000} and Name\orcidAuthor{0000-0000-0000-0000}}
    \affiliation{}
    \email{\{name\}@mail.com}
    }
\else
    % \author{\authorname{Name \orcidAuthor{0000-0000-0000-0000},Name\orcidAuthor{0000-0000-0000-0000} and Name\orcidAuthor{0000-0000-0000-0000}}
    % \affiliation{}
    % \email{\{name\}@mail}
    % }
    \author{\authorname{Thorsten Hempel\orcidAuthor{0000-0002-3621-7194}, Laslo Dinges\orcidAuthor{0000-0003-0364-8067} and Ayoub Al-Hamadi\orcidAuthor{0000-0002-3632-2402}}
    \affiliation{Neuro-Information Technology, Faculty of Electrical Engineering and Information Technology,\\ Otto von Guericke University, Magdeburg, Germany}
    \email{\{thorsten.hempel, laslo.dinges, ayoub.al-hamadi\}@ovgu.de}}
\fi


\abstract{Emotion expressions serve as important communicative signals and are crucial cues in intuitive interactions between humans. Hence, it is essential to include these fundamentals in robotic behavior strategies when interacting with humans to promote mutual understanding and to reduce misjudgements. We tackle this challenge by detecting and using the emotional state and attention for a sentiment analysis of potential human interaction partners to select well-adjusted engagement strategies. This way, we pave the way for more intuitive human-robot interactions, as the robot's action conforms to the person's mood and expectation. We propose four different engagement strategies with implicit and explicit communication techniques that we implement on a mobile robot platform for initial experiments.  }

\onecolumn \maketitle \normalsize \setcounter{footnote}{0} \vfill

\section{Introduction}
Once utopian, robots are increasingly moving from industrial and laboratory settings to the real-world in order to assist humans in everyday tasks. In this process, human-robot interaction became a central point of interest in robotic research that investigates the manifold challenges in performing interactive and collaborative tasks. As a general principle, the robot needs to be acquired with the necessary skills to enable intuitive interactions with arbitrary human interaction partners, regardless of the corresponding task, the human's intention and communication behavior. To this end, a fundamental objective is the anticipation of appropriate strategies to proactively approach or evade people based on the situation-specific context, such as the person's mood, attitude~\cite{Mul2019,Elp16} and ressentiments towards robots~\cite{Naneva2020ASR,10.3389/fpsyg.2021.592711}. 
First, it requires the determination of people's interest in interacting at all~\cite{6256006,1513748}. Thereupon, the robot has to find an appropriate strategy for either engaging people to establish an interaction or to avoid a confrontation if desired. In order to reach these underlying decisions, the robot has to cope with a row of sub-tasks: estimating people's focus of attention~\cite{8206301}, predicting their mental state,
and carrying out a (dis-)engagement strategy~\cite{Avelino2021BreakTI}.

Especially, the estimation of the interaction willingness is a very challenging task, as this mental state tends to provide only vague social signals. As an alternative approach, many methods~\cite{10.3389/frobt.2020.00092} go after the detection of engagement in order to determine if a person already approached the robot and waits for the robot's reaction. This state expresses itself more clearly, e.g., in voice commands~\cite{foster}, gestures and proxemics, but it undermines the proactive approach and neglects situations where the human partner is in need of help, but too uncertain or not aware of the robot. 

In this work, we close the gap between the lack of proactive behavior and advanced mental state predictions by introducing a sentiment-based engagement strategy.

By combining the focus of attention and emotions, we derive a sentiment analysis that allows us to select fine-grained behavior patterns that exceed current binary engagement and disengagement strategies. We implement our approach on a mobile robot platform and execute our strategies using explicit and implicit embodied communication. Our models are trained on minimum sized networks in order to perform our method on mobile systems with limited computational resources.

\section{Related Work}
In recent years, proxemic rules have been a popular tool to draw conclusions about the different states of human-robot interactions~\cite{6281352} and how spatial zones can be leverages to improve them~\cite{4415252}. \cite{8594149} predict appropriate encounter points to achieve a natural engagement with a group of people. Similarly, \cite{6256006} proactively approach detected people to offer help. \cite{10.1145/3434074.3447205} verbally greet people and dynamically adapt the voice volume based on the distance to the target person. Following \cite{Kendon1990}, \cite{Carvalho2021HumanRobotGT} apply the Kendon's greeting model to approach people while tracking the human mental states of the interaction using multiple features, such as gaze, facial expressions, and gestures.
Yet, all of these methods neglect if the human counterpart is actually interested and ready to initiate an interaction. This is problematic, as in the case of negative attitude, the robot's interaction efforts can be perceived as rude and annoying~\cite{Bro19}.
An initial attempt to address this was presented by \cite{8520646}, who only approach and offer help if they detect signs of attention towards the robot. But this still doesn't incorporate the actual mental attitude of the human counterpart towards the interaction itself. 

The recent advances in the area of deep-learning opened up new possibilities for the visual analysis of humans, starting from general detection tasks to the estimation of facial micro expressions. This allows the aggregation of additional relevant information, such as emotions~\cite{CHUAH2021102551,10.3389/frobt.2020.532279}, to better understand the behavior and intentions of the human interaction partner. To the best of our knowledge, we are the first to leverage this advancement by predicting sentiment states based on visual emotion estimation to improve interaction strategy for robots.
We summarize our main contributions as follows:
\begin{itemize}
    \item We propose an image-based sentiment analysis to gauge the current interaction preference of potential human interaction partners.
    \item We propose four different behavior strategies (Engage, Attract, Ignore, Avoid) that comprises a number of different explicit and implicit communication modalities.
    \item We design ultra lightweight models to implement our method and integrate it on a mobile robot platform.  
    \item We conduct initial experiments in laboratory settings for an early evaluation.
\end{itemize}

\section{Method}
This section explains each step of our proposed method for sentiment-based interaction behavior. 

\begin{figure*}[t]
\centering
\includegraphics[width=\linewidth]{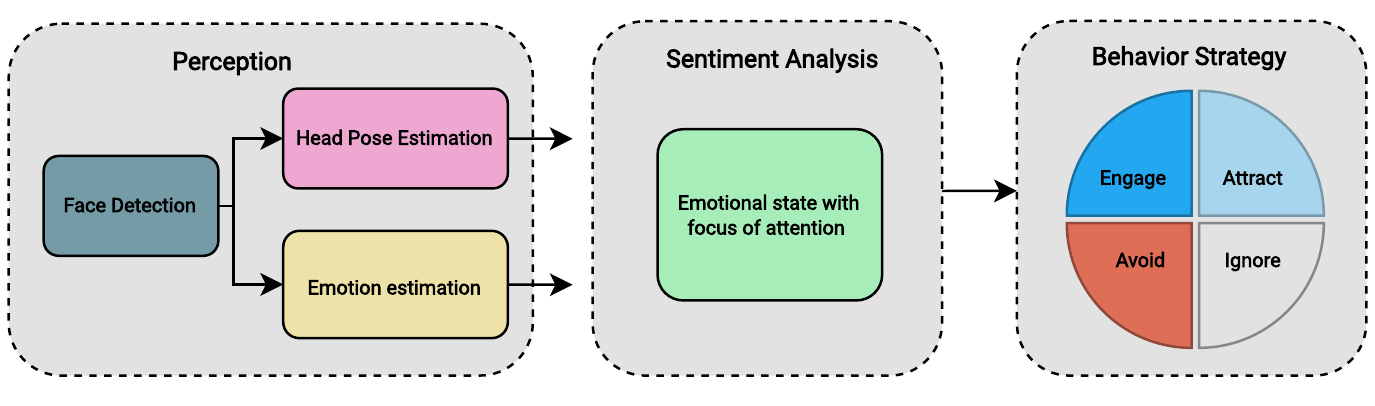}
\caption{Method overview of our proposed system. Detected faces are used to estimate the head pose and the corresponding emotions based on facial expressions. Both information is used to determine the emotional state with the corresponding visual focus of the person's attention that indicates the sentiment and allows a fine-tuned behavior strategy. }
\label{overview}
\end{figure*}

The sentiment analysis is based on two main features: head pose and emotion estimation. The head pose estimation is a leading indicator to gauge the current visual focus of attention of the person. The emotion estimation predicts the current emotional state of the person. Together, it allows the assumption about a person that  \textit{a)} is seeking for interaction, \textit{b)} is indecisive about it or \textit{c)} doesn't want to engage with the robot at all. Determining each sentiment state enables to perform a dedicated robot behavior that corresponds with the expected reaction from the person and, thus, improves an intuitive interaction. Figure \ref{overview} illustrates an overview of our proposed system. In the following subsection, we will give details about each component of the system and its interplay.

\subsection{Robot platform}
\begin{figure}[t]
\centering
\includegraphics[width=.9\linewidth]{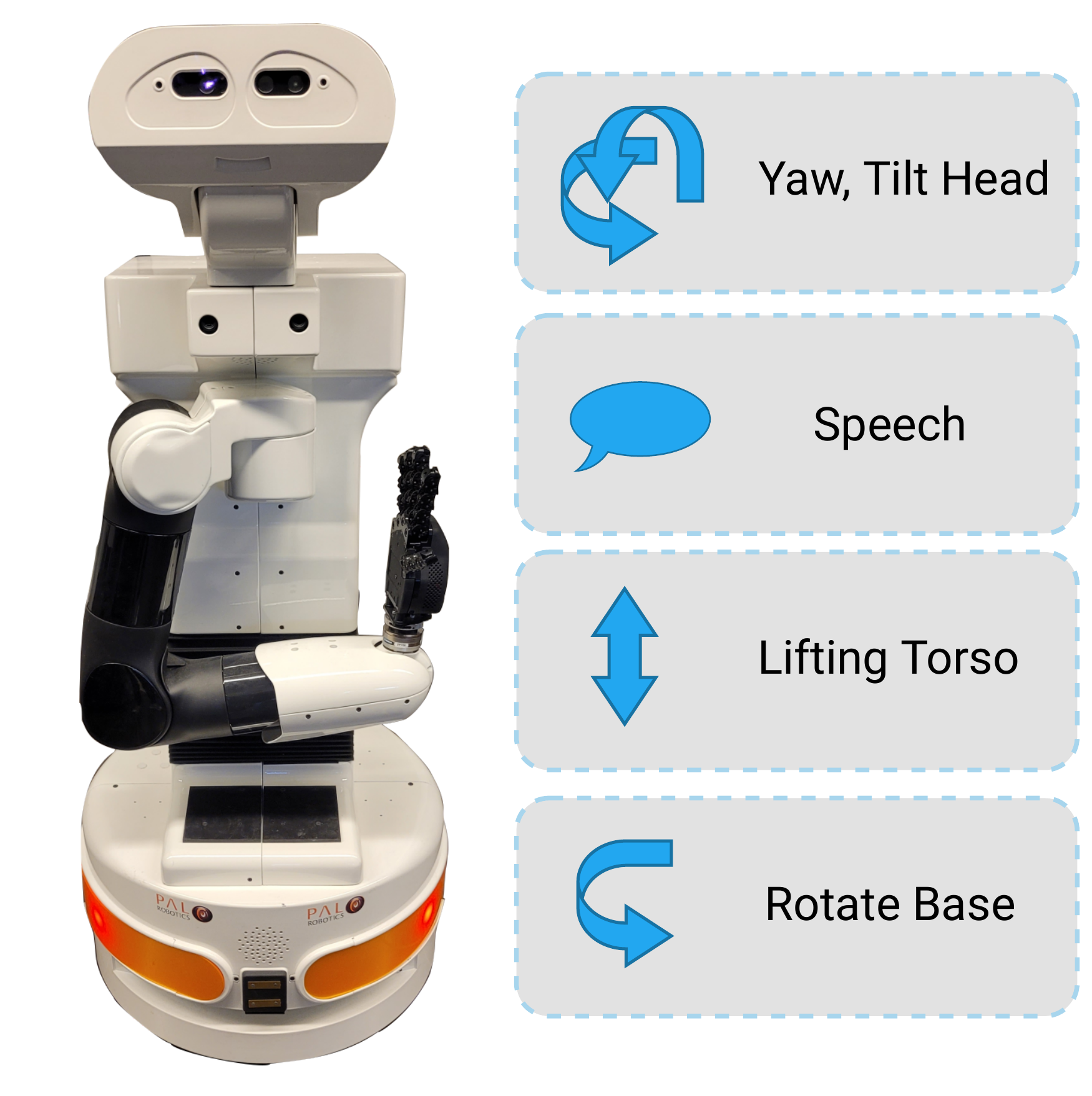}
\caption{The TIAGo robot platform with the key operations used in our method.}
\end{figure}
We use the TIAGo robot from PAL Robotics, a mobile service robot for indoor environments. It is equipped with an RGB-D camera mounted inside its head, that can be yawed and pitched to dynamically perceive the environment. For grasping and moving objects, the robot has a manipulator arm with a 5 finger gripper. The torso can be lifted to adjust the robot's height, and integrated speakers allow the output of voice commands. These abilities allow the combination of explicit and implicit communication modalities that we use in our method.
\begin{itemize}
    \item \textit{Head Following}: The robot's head aims at and follows a moving person as an implicit signal of attentiveness without actively approaching. 
    \item \textit{Body Following}: If a person exceeds the range of the robot's moving head, the entire robot base rotates in order to keep track of the person. This behavior communicates an attention commitment and is therefore an even stronger signal as the \textit{head following}.
    \item \textit{Torso Lifting}: Lifting the torso is a rather subtle communication cue, as it is perceived similarly to the social convention of standing up while greet someone. 
    \item \textit{Speech}: Approaching people with speech represents an explicit communication strategy, that stronger commits them to a reaction. Hence, speech should be  only  used if the approached person is highly expected to welcome an interaction.
\end{itemize}

\subsection{Visual Attention}
Typical indicators for gauging the human visual attention are the gaze and, more coarsely, the head pose. As the gaze depends on the eyes, which takes only a small part in face images, it is prone to errors. Therefore, we estimate the visual attention of the surrounding persons based on a head pose prediction algorithm.
At first, we locate faces in the image stream using an ultra light SSD face detector. Then, the face crops are further processed by the 6DRepNet~\cite{9897219}, that uses a rotation representation to directly regress yaw, pitch, and roll angle of the faces. To assure real-time processing capabilities, we replace the original 6DRepNet backbone with the efficient MobileNetv3-Small~\cite{Howard_2019_ICCV}. The new head pose prediction model is able to maintain 90\% of its accuracy compared to the original model, but is downsized to only one tenth of its parameters count.

\subsection{Emotion Estimation}
%facial expression as indicator for attention and attitude
In order to assess the attitude towards robots as well as differentiated subjective states such as dislike or willingness to cooperate, basic emotions can be used as indicating features. Naturally, these are frequently communicated by facial expressions. Hence, automatic facial expression recognition is an essential method to generate features for Human-Robot-Collaboration and attention prediction.

%End-to-end learning
Previous facial expression recognition methods adhere to the
established pipeline of face detection, landmark extraction,
and action unit (AU). Then, emotions are classified using these AUs as feature vectors \cite{Wegrzyn2017,Vin18,Werner2017}. However, end-to-end learning on a single but more comprehensive, and better generalizing database, often outperforms such traditional approaches. The AffectNet database~\cite{Mollahosseini2017AffectNetAD} for example contains about 500k in-the-wild samples for neutral and the basic emotions happy, disgust, fear, surprise, anger, and sadness. 
% valence & Arousal
Furthermore, it is the only database, that also includes ground truth of valence and arousal (VA), which are, in contrast to the basic emotion classes, continuous labels $\in \left[-1,1\right]$ which can be used for regression tasks.
VA is less intuitively interpretable by humans, but it allows differentiation between facial expressions of varying degrees. This is particularly useful for capturing modest but long-term changes in expressed emotion. In addition, a multitask network (simultaneous training of classification and regression) also improved the accuracy of the classification. Furthermore, Valence summarizes essential information about several emotions in one parameter (negative, neutral, positive), which may facilitate the development of heuristic behavior rules for some scenarios. 

%Model
\ifanonymous
Based on the work of \cite{Din21}, we implemented a multitask network. However, as a backbone for our multitask network,
\else
In our former work, we proposed a multitask network based on YOLOv3 \cite{Din21}. For the current paper,  
\fi
  we compared EfficientNetV2-S, ResNet18, and MobileNetV3-Large. Despite the poorer converging training loss, MobileNet, which is optimized for limited hardware, achieved almost the same accuracy on the test set as EfficientNet (deviation $< 1\%$). In addition, we used a weighted random sampler to handle the imbalanced dataset and re-train the network on the training and test set to generate an application model with increased robustness to unseen data.
\begin{figure}[t]
\includegraphics[width=\linewidth]{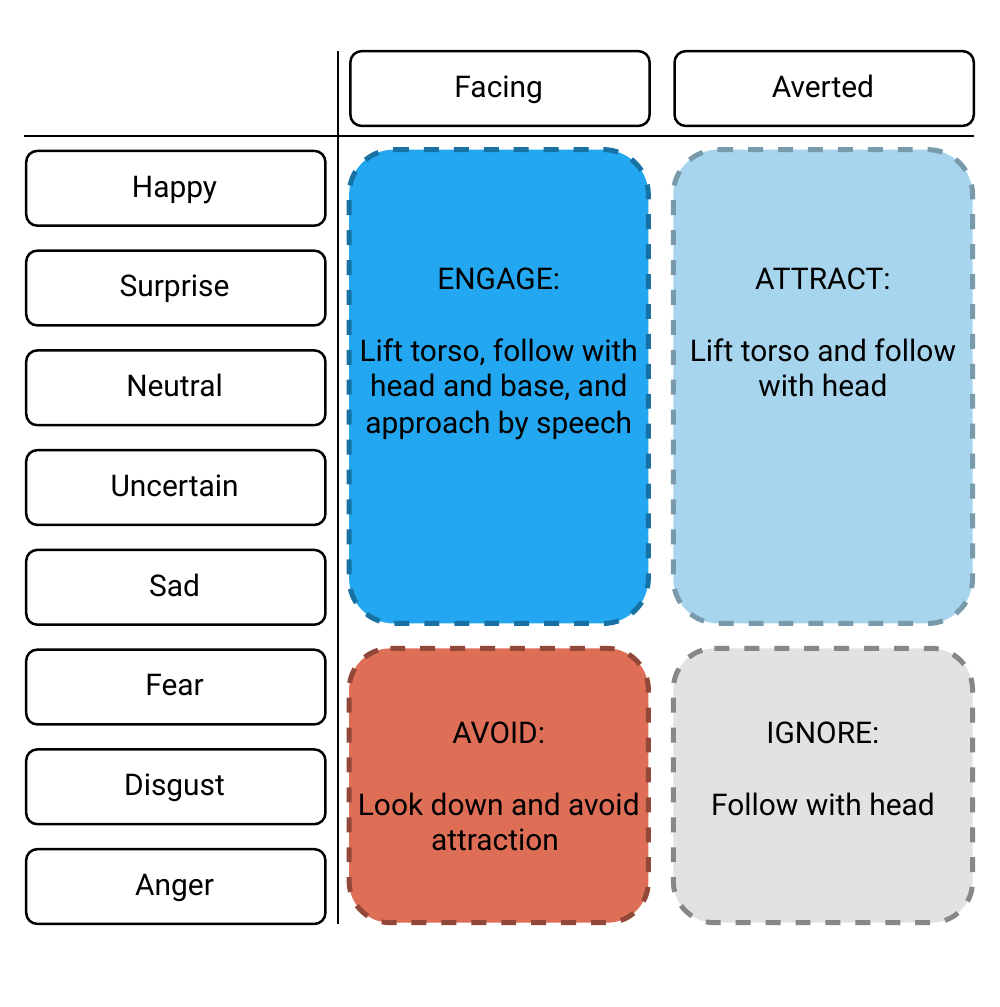}
\caption{Behavior strategies based on the emotional and attentive state.}
\label{fig:behavior-strategy}
\end{figure}
\subsection{Sentiment Analysis and Approaching Strategy}
Detecting emotional expressions combined with the visual focus of attention allows perceiving important communicative signals. As in human-human interaction, you may be more likely to greet a stranger who smiles at you but avoid eye contact if that stranger is scowling. We transfer these principles to determine states of sentiment based on the combination of the different emotions and visual attention. Is a person showing visual signs of fear or anger, while he is attentive towards the robot, it indicates a negative attitude towards the robot. In this case, a passive robot behavior without hectic movements can help to defuse the situation and to calm the person. If a person is feared, but not attentive towards the robot, the reason for the fear might not be the robot. Executing a more active robot behavior to show awareness could be therefore a more suitable strategy.
Hence, our sentiment analysis allows us to compartmentalize the common hard binary engagement strategy into soft behavior. Instead of choosing between engaging and not engaging, we choose between four different behaviors: Engage, Attract, Avoid, Ignore.
\begin{itemize}
    \item \textit{Engage}: Engaging will extend the torso of the robot to signal the awareness of the person's presence. Additionally, voice commands will greet the people to actively initiate an interaction. This strategy is called, when the attention is directed at the robot with clear positive emotional attitude.
    \item \textit{Attract}: Positive emotional attitude without attention towards the robot leads to more passive behavior. The goal is to signal the offer for interaction and help without actively committing to an interaction. This is done by extending the torso to show awareness, facing the person, but without verbally greeting. This way, a non-verbal offer for support (especially in case of uncertainty and sadness) is subtly given, that can be easily accepted or ignored.
    \item \textit{Avoid}: The avoiding strategy is executed when strong negative emotions (Fear, Disgust, Anger) are facing the robot. In this case, the robot avoids eye contact (keeps the face outside the camera center) and takes a passive part. Especially in case of fear, the goal is to increase the feeling of safeness by avoiding unexpected motions.
    \item \textit{Ignore}: If the negative emotions are not directed at the robot, the robot mostly ignores the person, only following it with its head to capture changes in emotional or attention states.
\end{itemize}
Figure \ref{fig:behavior-strategy} shows all combinations of emotional and attention states with corresponding behavior strategy.
 
 %https://viewer.diagrams.net/?tags=%7B%7D&highlight=0000ff&edit=_blank&layers=1&nav=1&title=Unbenanntes%20Diagramm.drawio#R7Zrvc5owHMb%2FGl62RxIVfEmt1r7YXsz1enuZQYRsgXAhVLu%2FfqChil93t9rZ1MHhcfDkh5APj4FHHDJJ13eK5sknGTHhYDdaO%2BTWwRgNMHbqjxs9bxXPd7dCrHhkKu2EBf%2FFjNhUK3nEilZFLaXQPG%2BLocwyFuqWRpWSq3a1pRTtb81pzICwCKmA6iOPdLJVB6670%2BeMx4k%2BKEhpU9cIRUIjudqTyNQhEyWl3m6l6wkT9dg1w7JtN%2FtD6ctxKZbpv2mwmN8j9Pj1B5qPouLqm58%2BTMdXppcnKkpzvnOa58%2FmiPVzMwpKllnE6p6QQ25WCddskdOwLl1V3Cst0akwxUuZ6RlNuaiRf5HfpZZGXchSbRolWlf88JAE1ao65HpVVyiuYyljwWjOi%2BtQppuCsNhUnS23fVabu17hKDSnxJRm6z3JjModkynTqurGXbevNHOBjszuakcbNQyTPdLEaNRcYPFLxzsI1Ybh8AomGDB5yMLqbCjPuswFDWyDIQDMolS54gXrNBfXNpcB4PKZlVpR0Wksvm0sQ2gXGnUZycutgDUkI4BkxqjqNBPr070HmNzyIi4L3WUsxPqk4gMsQRazTnuFWL8DG8PfLxryLO4WlcHBT5htKs1lsu%2BVzel0a7bH3kfjAp%2Fup1lc5x1vw8KFmEgh1aYtwZh6S7fSC63kT7ZXgjwPB%2BTyQB4azPotAoKRQKCrB5zwrfcIbZLUi4Z1U0jS88dBcHN5JA8taZ8kzBDOYsnAm%2F3XlrQfOiCYOvwbT74jmNf7%2Fwye%2FAAoYVLRz5OnmNL68wGCAUdvypNMaR8lzEV6U55iSus5MDoSpfSmPMWU9lHCAKY35SmRjfVsoPmy3pRvTXnso4QxT%2FAkqyaXCjJiIzYcHQM5RgOf%2BGfypPVwAMOY5z7OpLqg1wsOSDJcL8dI%2Bl69nMmS9knCmKe35Ostaf9fQAxTngu35BTXy3tb8gOQhCFPb8kTLGk9GMAw47lwS9qZJc9Jstrdvdi8Kdt7O5xMfwM%3D

\section{Implementation}
We used the ROS\footnote{https://www.ros.org/} middleware to implement our proposed solution on our robot platform, as this also offers a simplified integration into other ROS-based robot platforms. Face detection, head pose and emotion estimation are executed in separate nodes, that forward the necessary information to perform the sentiment analysis and the action controls. While the latter is performed on the robot itself, the inference of the neural networks is outsourced to a separate notebook that is mounted on the shoulder of the robot. This ensures the complete mobility of the robot, while exploiting additional computational resources. The notebook has a Quadro RTX 4000 processing all models in parallel with the following inference times:
\begin{itemize}
    \item Face Detection: \tab 6.7~ms
    \item Head Pose Estimation: \tab 1.4~ms
    \item Emotion Estimation: \tab 6.3~ms
\end{itemize}
The visual perception can be therefore performed in real time and is able to continuously track people in the surroundings.

\begin{figure}[t]
\includegraphics[width=\linewidth]{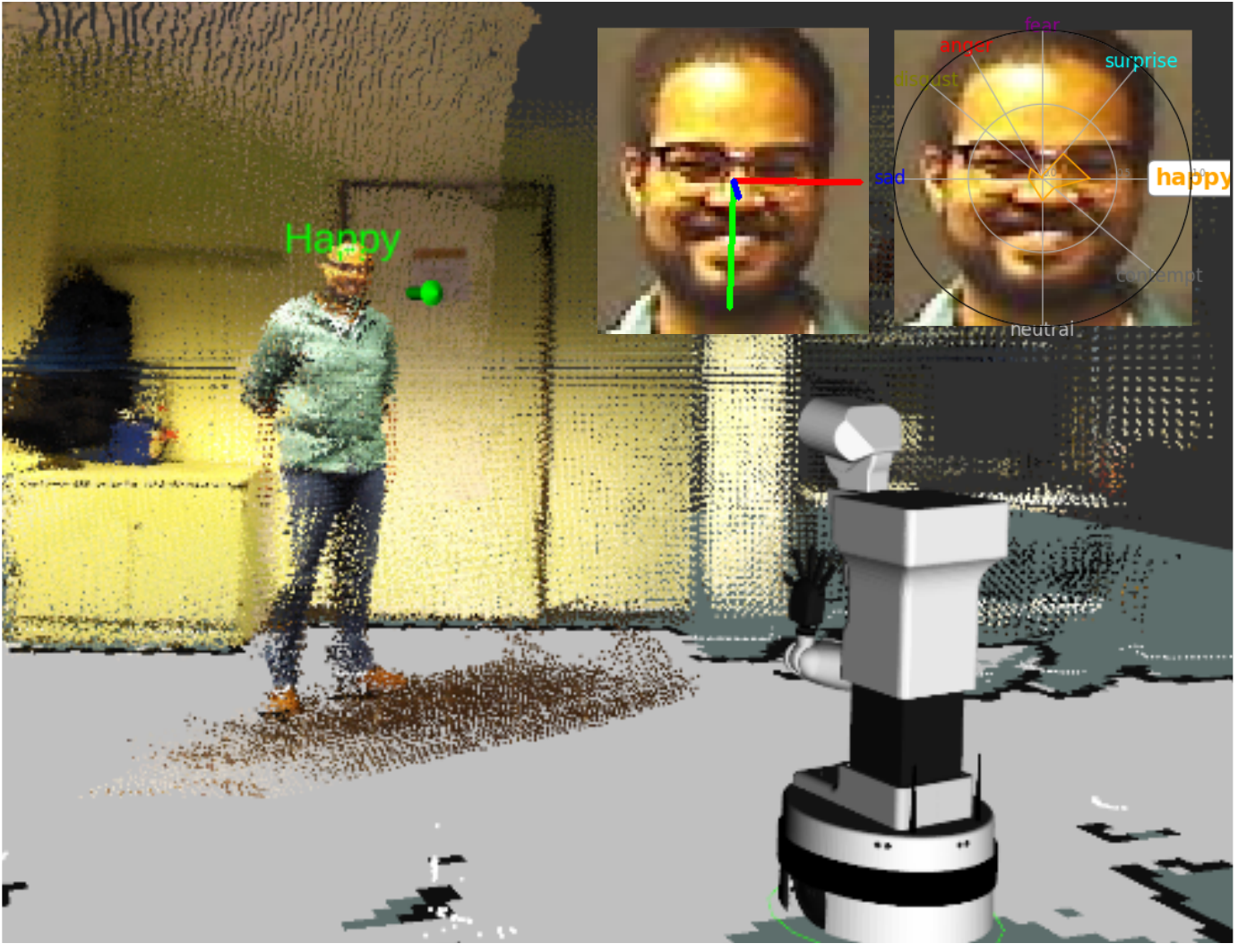}
\caption{Sentiment analysis in a laboratory setting. The person is tracked using a face detector, followed by a head pose and emotion estimation. As the person shows attentiveness and happy, a positive sentiment towards the robot is expected and a proactive initiation strategy selected.
}
\label{fig:experiment}
\end{figure}
\section{Experiment}
We conducted an initial experiment in a laboratory setting to test our implemented method on the TIAGo robot platform. The objective of the experiment was to gain experience about the model's performances in real application scenarios. Figure \ref{fig:experiment} shows a snapshot from one of our tests. The robot platform detects a face in the environment and predicts head pose and emotional state. The focus of attention is on the robot (indicated by the green arrow) in conjunction with expressing happiness, so the robot initiates an interaction by following the engage strategy. 

While the models provide very robust predictions, the implementation is currently entirely based on single shot detections. Yet, over the duration of face tracking, facial expressions change intermediately and short distractions make the head pose occasionally oscillating. This leads to abrupt changes of states within short time periods. Hence, to improve the robustness of the sentiment analysis, conclusions about the current state should incorporate the predictions from multiple time frames.

\section{Conclusion}
In this paper, we present a sentiment-based methodology to improve the intuitiveness and reasoning in human-robot interactions. Our method focuses on the visual perception of mental states and the focus of attention to derive a sentiment analysis for suitable interaction initiation strategies. We take also into account the case where people are not interested in an engagement at all, which has been mainly neglected in the recent literature.   
To implement our method, we trained lightweight perception models that we integrate on a mobile robot platform. Finally, we conducted initial experiments using a mobile robot platform to analyze the performance of our models and the overall system.

In the future, we will extend our work by embedding our sentiment analysis in a temporal-probabilistic framework to improve the robustness in real-world scenarios. Here, we will also incorporate the valence and arousal predictions to evaluate the emotion's intensity and transitions between emotional states. Further enhancements offer the inclusion of the robot's mobility to actively engage and disengage people or to add additional interaction modalities, such as following people by command, getting out of the way of passing people, or approaching a place that offers a better overview of the surroundings.

Finally, we will conduct a study with test subjects to evaluate our approach in terms of intuitiveness and sympathy. 

\vfill
\section*{\uppercase{Acknowledgements}}
%\textit{Removed for double-blind review}
This work is funded and supported by the Federal Ministry of Education and Research of Germany (BMBF) (AutoKoWAT-3DMAt under grant Nr. 13N16336) and German Research Foundation (DFG) under grants Al 638/13-1, Al 638/14-1 and Al 638/15-1.

\bibliographystyle{apalike}
{\small
\bibliography{bib,ref_emotion,ref_HRC}}

%\section*{\uppercase{Appendix}}

%If any, the appendix should appear directly after the
%references without numbering, and not on a new page. To do so please use the following command:
%\textit{$\backslash$section*\{APPENDIX\}}

\end{document}